\documentclass[lettersize,journal]{IEEEtran}
\usepackage{amsmath,amsfonts}
\usepackage{algorithmic}
\usepackage{algorithm}
\usepackage{array}
\usepackage{color}
\usepackage[caption=false,font=normalsize,labelfont=sf,textfont=sf]{subfig}
\usepackage{textcomp}
\usepackage{stfloats}
\usepackage{url}
\usepackage{verbatim}
\usepackage{graphicx}
\usepackage{cite}
\usepackage{booktabs}
\usepackage[table]{xcolor}
\usepackage{todonotes}
\newcommand{\xcont}{\mathbf{X}_{cont}}
\newcommand{\xint}{\mathbf{X}_{int}}
\newcommand{\xcat}{\mathbf{X}_{cat}}

\begin{document}

\title{Exploratory Landscape Analysis for\\Mixed-Variable Problems}


\author{
Raphael Patrick Prager and 
Heike Trautmann\\

\thanks{Manuscript received September 24, 2023. 
}
\thanks{
Raphael Patrick Prager is with the Department of Information Systems, University of M\"unster, Leonardo-Campus 3, 48149 M\"unster, Germany.\\
E-mail: raphael.prager@uni-muenster.de}

\thanks{
Heike Trautmann is with the Computer Science Department, Paderborn University, Fürstenallee 11, 33102 Paderborn, Germany and with the Data Management and Biometrics Group at the University of Twente, NL\\
E-mail: heike.trautmann@uni-paderborn.de}
}

\markboth{Transactions on Evolutionary Computation,~Vol.~x, No.~x, September~2023}%
{Prager \MakeLowercase{\textit{et al.}}: Exploratory Landscape Analysis for Mixed-Variable Problems}


\maketitle

\begin{abstract}
Exploratory landscape analysis and fitness landscape analysis in general have been pivotal in facilitating problem understanding, algorithm design and endeavors such as automated algorithm selection and configuration. These techniques have largely been limited to search spaces of a single domain.
In this work, we provide the means to compute exploratory landscape features for mixed-variable problems where the decision space is a mixture of continuous, binary, integer, and categorical variables.

This is achieved by utilizing existing encoding techniques originating from machine learning. We provide a comprehensive juxtaposition of the results based on these different techniques. 
To further highlight their merit for practical applications, we design and conduct an automated algorithm selection study based on a hyperparameter optimization benchmark suite. 

We derive a meaningful compartmentalization of these benchmark problems by clustering based on the used landscape features. The identified clusters mimic the behavior the used algorithms exhibit. Meaning, the different clusters have different best performing algorithms. 
Finally, our trained algorithm selector is able to close the gap between the single best and the virtual best solver by $57.5\%$ over all benchmark problems. 
 
\end{abstract}

\begin{IEEEkeywords}
Exploratory landscape analysis, mixed-variable problem, mixed search spaces, automated algorithm selection.
\end{IEEEkeywords}

\section{Introduction}
\IEEEPARstart{E}{xploratory} landscape analysis (ELA)~\cite{Mersmann2011} and similar techniques to numerically characterize fitness landscapes have been present for the last two decades. These methods have enabled and facilitated research in fields such as automated algorithm selection (AAS) and configuration~\cite{kerschke2018tsp, KerschkeT2019AutomatedAlgorithm, KerschkeHNT2019AutomatedAlgorithm}. Prominent examples are fitness distance correlation metrics~\cite{jones95fdc} or local optima networks~\cite{ochoa08lon} for the combinatorial optimization domain, whereas ELA provides the means for the continuous problem space. However, the different problem domains are not intrinsically disjoint. In fact, many contemporary optimization problems consist of a combination of continuous and discrete decision variables. A first attempt to enable ELA for mixed-integer problems has been undertaken by our work in \cite{prager23mip}. There, we demonstrate that ELA features, originally designed for the continuous space, can be applied meaningfully in mixed-integer spaces without requiring any modifications to the existing implementation. Another attempt to extend the usage of ELA to the purely binary domain was undertaken by \cite{wmodel} with promising results. While certainly a step in the right direction, many real-world problems often include a variety of different decision variables including variables which lack any ordinal structure. Meaning, these problems are constituted by a mix of continuous, integer, and categorical decision variables. The complexity is heightened by the presence of conditional/hierarchical structures where certain decision variables are only of relevance when another decision variable attains a certain value.

This complex type of problems is referred to as mixed-variable problems (MVP)~\cite{pelamatti18mvp}. Only few works have discussed and developed means to analyze fitness landscape of MVPs. In~\cite{pim20fdcautoml}, the authors used the aforementioned fitness distance correlation measures to gain insights in the landscape of automated machine learning (AutoML) landscapes. This was made possible by encoding the different candidate solutions as tree structures and utilizing an appropriate distance metric. The drawn conclusions point to a poor fit of this feature set, at least for AutoML landscapes. Another method was proposed by~\cite{push18ac} and is called parameter response slices. In that and a subsequent work~\cite{push22loss}, the authors identified that the majority of considered AutoML problems are unimodal and highly structured. While these provide valuable insights, by the authors' own admission these methods are prohibitively costly. This renders them unsuitable for areas such as AAS.

Hence, our intent is to enable the existing set of landscape features of ELA for meaningful usage on MVPs. 
The rationale behind this is that a majority of ELA features are inherently cheaper to calculate and therefore applicable for AAS compared to~\cite{push22loss}.
We illustrate the merits of our devised approach on a set of hyperparameter optimization problems~\cite{bischl23hpo}, wherein we conduct a comprehensive analysis of the resulting ELA features. In addition, we construct an AAS setting in which we utilize a set of complementary solvers and the adapted ELA features to automatically select an appropriate algorithm for a given problem. This endeavor has shown to be effective as it surpasses the performance of the single best solver within our  algorithm portfolio. Consequently, we can ascribe our formulated approach and the subsequent ELA features a substantial degree of discriminative power. At this point, we want to emphasize that the purpose of this work is not to derive insights for hyperparameter optimization but for MVPs in general. The chosen set of problems is simply a means to an end as it encapsulates a variety of challenges often in encountered in mixed-variable optimization.

This paper is organized as follows. The formal definition of MVPs is presented in Section~\ref{sec:mvp}, followed by a succinct description of the used ELA features in Section~\ref{sec:ela}. In Section~\ref{sec:challenges}, we detail the obstacles which prohibit the usage of ELA for MVPs and provide a solution-oriented approach, respectively. This is followed by Section~\ref{sec:exp-setting} in which we state the different experimental components necessary for an AAS scenario. These are the set of benchmark functions, the algorithm portfolio and performance, the generation of ELA features, and the training of the algorithm selection model. For each component, we provide an analysis and report interesting results. Finally, in Section~\ref{sec:conclusion}, we conclude this paper and contextualize our findings around current limitations and possible avenues for future research.

\section{Mixed-Variable Problems}\label{sec:mvp}
In the context of MVPs, we can distinguish between two types of variables: a set of continuous variables denoted as $\xcont$ and a set of discrete variables. The latter can be further subdivided into variables with an ordinal structure ($\xint$) and without an ordinal structure ($\xcat$), which are also known as categorical variables. Although elements in $\xcat$ assume no natural ordering relation, they are commonly represented by integer values \cite{pelamatti18mvp}.

One of the major challenges faced by many MVPs is that a certain value of a discrete decision variable influences the dimensionality and subsequent constraints of the problem itself. In other words, the feasible domain of $\xcont$ may be determined by values of $\xint$ and/or $\xcat$.

Dependent on the research domain, these variables are referred to as `dimensional', `hierarchical', `conditional', or `nested' variables.
For engineering problems, this can occur when one variable allows to select between different types of equipment where each of which possesses its own set of parameters~\cite{pelamatti18mvp}. A different example from hyperparameter tuning can be related to a support vector machine, where the choice of a particular kernel introduces additional parameters to tune~\cite{bischl23hpo}. We adopt the terminology of the used benchmark suite (cf. Section~\ref{sec:exp-setting}) and will refer to this phenomenon as hierarchical. 

In this study, we decide to relax these constraints. This choice was made because it offers the added benefit that the numerous ELA features, proposed throughout recent decades, can be generalized to MVPs whereas the alternative would be the excruciating and time consuming development of new features tailored to capturing these hierarchical structures. We deem it more auspicious to start with an existing set of features and exploit the potential of MVP specific landscape features in future research. Further elaboration on this topic is provided in Section~\ref{sec:challenges}. It should be noted that we do not discard any problem instance with hierarchical structures, only that we relax the constraints and examine the effect this has on ELA features in comparison to problem instances that lack hierarchical structures.

With the previously defined vectors, we can define the decision variables of our MVP as $\mathbf{X} = (\xcont, \xint, \xcat)$.
The MVP itself is formalized w.l.o.g in a similar manner as described in \cite{lucidi05mvp2, pelamatti18mvp}:
\begin{align}\label{eq:mvp}
\begin{split}
    \text{min}\quad & f(\xcont, \xint, \xcat) \\
    \text{w.r.t.}\quad & \xcont \in  \mathbb{R}^{n_{cont}} \\ 
    & \xint \in \mathbb{Z}^{n_{int}} \\
    & \xcat \in \mathbb{Z}^{n_{cat}} \\
    \text{s.t.}\quad & \mathbf{g}(\xcont, \xint, \xcat) = 0 \\
    & \mathbf{h}(\xcont, \xint, \xcat) \leq 0
\end{split}
\end{align}

Here, $f$ represents the single-objective function, with equality constraints $\mathbf{g}$ and inequality constraints $\mathbf{h}$. Note that this formal definition of MVPs does not account for hierarchical structures.

\section{Exploratory Landscape Analysis}\label{sec:ela}
In black-box optimization, the fitness landscape of any problem instance cannot be analyzed analytically as there is no closed-form mathematical representation of it.
In the domain of single-objective \textit{continuous} optimization, ELA has been developed as a remedy. ELA provides the means to quantify certain structural characteristics of fitness landscapes. These structural characteristics have been termed `high-level properties' by \cite{mersmann10properties}. In a later work, so-called `low-level feature sets' have been developed to map these low-level feature sets to certain high-level properties \cite{Mersmann2011}.

These low-level feature sets are based on a well-distributed sample $S$ of the decision space and the corresponding objective values $Y$. Hence, this sample is defined as $S = (\mathbf{X}, Y)$ and often referred to as \textit{initial design}.

This initial design is used to generate summary statistics and distance-based calculation to provide different sets of low-level features. These low-level feature sets have been used in numerous studies for AAS~\cite{PragerTWBK2020PerInstance, KerschkeT2019AutomatedAlgorithm, KerschkeHNT2019AutomatedAlgorithm} and general studies pertaining fitness landscapes~\cite{ochoa08lon}. Later studies have further augmented the set of available low-level feature sets. Two packages, namely \texttt{flacco}~\cite{KerschkeT2019flacco} and \texttt{pflacco}~\cite{pflacco}, provide an implementation of these features in the programming languages R and Python respectively.

In this work, we use a subset of these ELA features which have been predominantly used in the research community and shown to be highly effective for AAS. A succinct description of this subset is given in the following:

 \begin{itemize}
     \item \textbf{Classical ELA} (12 features): 
     Under the term `classical ELA' features fall six feature sets which were designed by~\cite{Mersmann2011}. From this amalgamation of feature sets only two are of particular interest in this study. These are \texttt{ela\_distr} and \texttt{ela\_meta}. The former calculates the kurtosis and skewness of objective values. The latter constructs several linear and quadratic models and thereby measures the degree of linearity and convexity of a fitness landscape. 
     We remove the other four feature sets from consideration because they either proved to be less significant in previous studies or require additional expenditure of function evaluations beyond the initial design~\cite{Mersmann2011}.

     \item \textbf{Dispersion} (16 features): Based on different levels of quantiles (e.g. 10\%, 25\%), the dispersion feature set compartmentalizes the objective values of the initial design into several subsets. For each subset, the mean and median distances in the decision space within a group is measured. This gives insight into whether local optima are localized in a small region or are distributed throughout the landscape~\cite{lunacek2006}.
     
     \item \textbf{Information Content} (5 features):
     The calculation of this feature set relies on a series of random walks conducted over the initial design. In each iteration, a comparison is made between the current observation and the subsequent one. From this, various metrics are generated to characterize the landscape in terms of its smoothness, ruggedness, and neutrality~\cite{munoz2015_ic}.
     
     \item \textbf{Nearest Better Clustering} (5 features): 
     For each observation in the initial design, the distance to its nearest neighbor and nearest better neighbor are calculated. The term `better' refers to a qualitative improvement of the objective value. For each of these two sets of distances, certain metrics and ratios are derived and compared to each other~\cite{funnelkerschke2015}.
 \end{itemize}


\section{Preprocessing Steps for Exploratory Landscape Analysis}\label{sec:challenges}

\begin{figure}[ht!]
    \centering
    \includegraphics[width=\columnwidth]{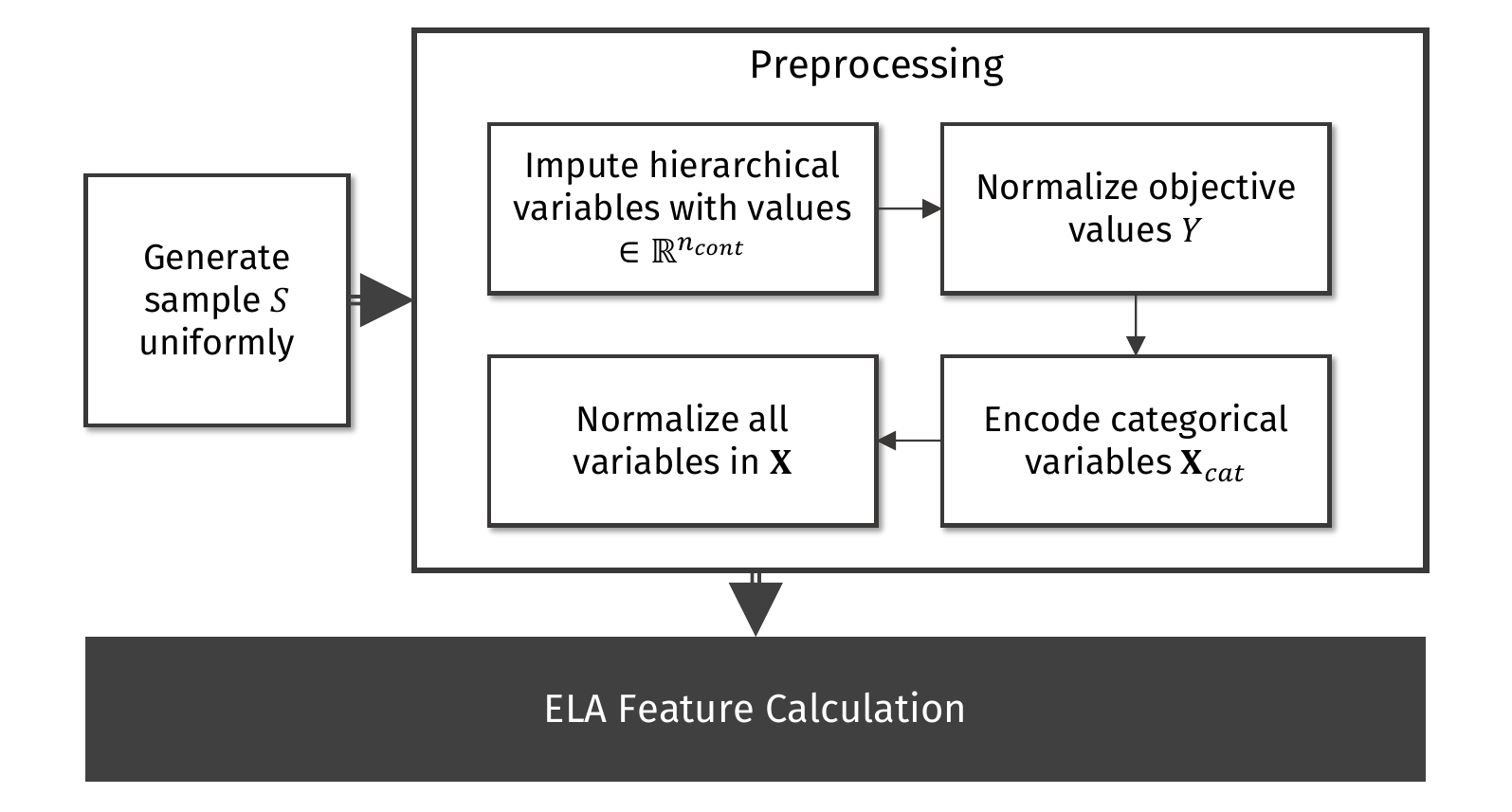}
    \caption{High-level overview of the different steps prior to ELA feature computation for MVP.}
    \label{fig:workflow}
\end{figure}
In this section, we enumerate and discuss the individual preprocessing steps that we undertake to enable the calculation of ELA features for MVPs. A general overview of the existing processing steps is given in Fig.~\ref{fig:workflow}. 

Opposed to previous work conducted in the continuous domain, we generate our sample uniformly at random. The rationale for this is the scarcity of research on space-filling designs for the mixed-variable spaces. As this preprocessing step is rudimentary, it does not warrant further elaboration.
The four remaining steps are detailed in the following subsections.

\subsection{Hierarchical Decision Space}\label{sec:challenges-hier}

\begin{figure}[!t]
    \centering
    \includegraphics[width=0.75\columnwidth]{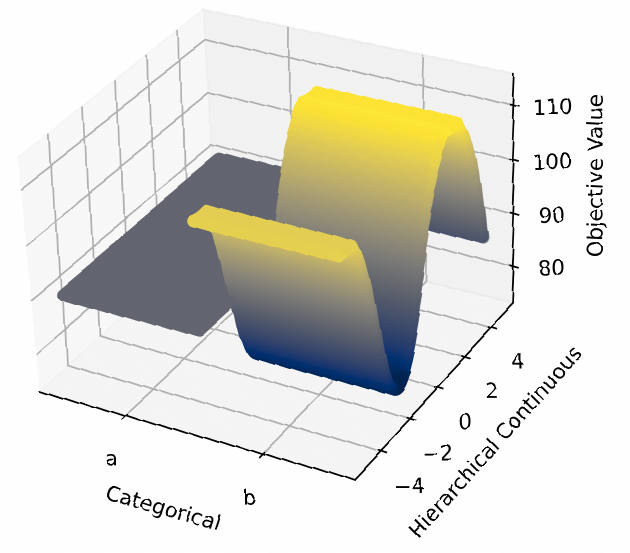}
    \caption{Exemplary hierarchical fitness landscape. The domain of $X_{cat}$ comprises the value a and b. $X_{cont}$ only affects the fitness landscape when $X_{cat} = b$.}
    \label{fig:cond_landscape}
\end{figure}
Hierarchical decision spaces pose a significant challenge in general. When observed from an isolated landscape perspective, these hierarchical features can be viewed as any other decision variable where we essentially relax the constraints imposed by the hierarchical structure. Thereby, infeasible solutions become feasible. When viewing the fitness landscape, these formerly infeasible areas are now large plateau shaped regions with a constant objective value. This information can be incorporated into existing ELA features as this is also an occurrence in continuous optimization problems.

To illustrate this, let us assume that a particular MVP exists with an objective function $f'(X_{cont}, X_{cat}) \in \mathbb{R}$. The search space consists of two decision variables. The first decision variable is categorical with $X_{cat} \in \{a, b\}$ and the second is continuous with $X_{cont} \in [-5, 5]$. In addition, $X_{cont}$ is dependent on the value of $X_{cat}$, i.e., it only influences $f'$ when $X_{cat} = b$. An exemplary fitness landscape with an arbitrary definition of $f'$ and the aforementioned restrictions is depicted in Fig.~\ref{fig:cond_landscape}.
A large area of the fitness landscape is constituted by a plateau where the hierarchical decision variable $X_{cont}$ does not have any effects on the landscape. This occurs when $X_{cat}$ attains the value $a$. The subspace spanned by $f'(X_{cat} = b, X_{cont})$ depicts a stark contrast. Here, small changes in $X_{cont}$ strongly affect the quality of the found solution. As it is the case in many MVP or mixed-integer problems (even without hierarchical structures) the fitness landscape is highly discontinuous.
However, capturing these plateau shaped subspaces in MVPs provides meaningful information for landscape analysis in general.

While we relaxed the conditions for our MVP conceptually, the MVP itself still adheres to these constraints and does not allow infeasible solutions. In order to evaluate a candidate solution and simultaneously adhere to the constraints, a candidate solution might be composed of empty entries for hierarchical decision variables. These empty entries are usually represented by \texttt{NA} values. In the context of the previous example, this means that the MVP expects a candidate solution in the format $f'(a, \texttt{NA})$ and not as $f'(a, 3.5)$.

Hence, after generating the sample $S$ (including the objective values $Y$), we impute the respective \texttt{NA} values with random values (e.g., 3.5) of the variable's respective domain. We want to emphasize again that we only relax the constraints of the hierarchical structure to calculate ELA features. This does not apply to any other subsequent evaluations of these MVPs, e.g., by benchmarking algorithms.

\subsection{Different Ranges in the Search Space}\label{sec:challenges-ranges}
Another aspect which warrants contemplation is the potentially varying scale of decision variables, which is often encountered outside of artificial benchmark problems. For example, it could be the case that one decision variable allows for continuous values in the interval $[0, 1]$ whereas a second variable has values in the interval $[-5, 5]$.
For many previous works, this has not been a concern because they were performed on the equally sized box-constraint black-box optimization benchmark (known as BBOB)~\cite{Hansen2009}. However, in a previous study we already demonstrated the detrimental effects of different ranges in the objective space for \textit{some} ELA features~\cite{prager23nully}. This also applies to some extent to the decision space. Some features are distance-based and are thereby greatly influenced by the range of possible values in the decision space. For a single problem instance, this already may lead to problems as differences in decision variables with large ranges will artificially effect the respective ELA features more. This issue is further exacerbated when we try to compare ELA features between \textit{different} problem instances.

Therefore, we apply min-max normalization to the entire initial design prior to the ELA feature computation. We use the known minima and maxima of the respective decision variables as these are typically constrained. The objective space on the other hand is normalized by using the respective sample minimum and maximum as suggested in~\cite{prager23nully}. 

Note that this step occurs at different points in the preprocessing scheme. The normalization of the objective values takes places before we transform categorical variables to a numerical representation. The reason for this is that some encoding variants are dependent on the actual objective values of $Y$.
After this transformation, the entire decision space $\mathbf{X}$ is normalized (cf. Fig.~\ref{fig:workflow}).

\subsection{Transformation of Categorical Decision Variables}\label{sec:challenges-enc}
\textbf{One-Hot Encoding (OH)}. The challenges imposed by categorical variables, which lack any ordinal structure, have been persistent issues for several decades. One simple yet effective approach to deal with categorical variables is to employ one-hot encoding (OH). To this date, OH continues to perform astonishingly well and is widely adapted in the ML community~\cite{Hancock2020}.

Let $V$ be a set of non-numeric values $V = \{v_1, v_2, ..., v_n\}$. with a decision variable $X_{cat} \in V$. Then, OH is a defined as:
\begin{align}
\begin{split}
    t_{X_{cat}}:\quad & V \to \mathbb{Z}_2^{n} \\
    \text{with}\quad &t_{X_{cat}}(v_k) = (z_1, \ldots, z_k, \ldots, z_n)\\
    \text{s.t.}\quad &z_k = 1 \\ 
    & z_{k'} = 0 \quad \forall k' \in \{1, ..., n\}\setminus\{k\}.
\end{split}
\end{align}
%


For a more concrete example, let $x_1$, $x_2$, and $x_3$ be a realization of a single decision variable $X_{cat}$ with $x_1 = a$, $x_2 = b$, and $x_3 = c$. A possible transformation might look like this $t(x_1) = (1, 0, 0)$, $t(x_2) = (0, 1, 0)$, and $t(x_3) = (0, 0, 1)$.

Although not a concern in the given example, the effectiveness of OH diminishes as the number of of high-cardinality categorical variables increases. For instance, if we consider a dataset consisting of three random variables, where each has a cardinality of $10$, the resulting dataset would be transformed into a $30$-dimensional dataset. This forces additional complexity onto methods which operate on the transformed data and introduces other challenges as, for example, the curse-of-dimensionality for distance-based calculations. Moreover, it can distort the correlation between decision variables and objective values by the introduction of the new encoding variables.

\textbf{Target Encoding (TE)}. Developed by \cite{te2001}, target encoding (TE) is a preprocessing technique originating from the area of supervised learning. At its core, it transforms a categorical variable of arbitrary cardinality to a single scalar. Thereby, the dimensionality of the original dataset remains the same (in contrast to OH). In general, supervised learning can deal with classification or regression tasks.
For classification tasks, TE uses Bayesian probability models to estimate the probability of each class for each category of a given categorical decision variable. In regression tasks, the influence of a certain category on the expected value of the target value is measured.

In our case, the target is not the typical label or scalar of a classification or regression task, but the objective values $Y$ of an MVP. Based on the initial sample $S = (\mathbf{X}, Y)$, the transformation of a specific category $j$ of a variable $X_{cat} \in \mathbf{X}$ is defined as follows:
\begin{equation}\label{eq:te}
j' = \lambda(n_j) \frac{\sum_{i \in Y_j}y_i}{n_j} + (1 - \lambda(n_j)) \frac{\sum_{i = 1}^{n}y_i}{n}, 
\end{equation}

where $\lambda$ is a smoothing function producing values between $[0, 1]$, $Y_j$ is a subset of $Y$ of size $n_j$ which only contains observations where the corresponding $X_{cat}$ variable attains the category $j$ in question. This transformation replaces the category $j$ in variable $X_{cat}$ with the newly calculated value $j'$.

The purpose of the function $\lambda$ is to balance the uncertainty of the estimated expected value. When the category $j$ is not sufficiently represented in $X_{cat}$ the resulting estimated is primarily based on the latter half of Equation~\ref{eq:te}, i.e., the arithmetic mean of $Y$.

In \cite{te2001}, two different smoothing functions are proposed, where one offers an additional external parameter. We opted for the other (Equation 8 in \cite{te2001}) as tuning another parameter would add an additional layer of complexity.
The used smoothing function is defined as:
\begin{equation}
    \lambda(n_j) = \frac{n_j}{n_j + \frac{\sigma^2_j}{\rho^2}},
\end{equation}
where $n_j$ remains the same as before, $\sigma^2_j$ is the variance of $Y_j$ and $\rho^2$ is the variance of $Y$.


\section{Experimental Methods}\label{sec:exp-setting}
A purely theoretical analysis at this moment is severely limited. Mainly, because we are lacking information about the high-level properties of the problem landscapes in question. For instance, it is difficult to visually or theoretically asses whether a landscape of an MVP is multi-modal or not. Although, \cite{push22loss} speculate, not without substance, that AutoML landscapes are rather benign and well structured. However, this is restricted to only a subset of MVPs. 
Hence, we experimentally determine the usefulness of categorical ELA features under various lenses. From there, we plan to infer any theoretical implications in future works.

For this work, that means that we apply our preprocessing approach, including the two transformation variants of categorical decision variables, in the context of AAS. For any AAS model, we require a set of problem instances, a collection of algorithms which complement each other in terms of performance, a set of landscape features which numerically characterize the problem instances, and a machine learning (ML) model to learn a mapping which automatically assigns an algorithm to a given problem instance.

\subsection{Benchmark Functions}
The used set of benchmark functions is provided by `Yet Another Hyperparameter Optimization Gym' (YAHPO Gym)~\cite{yahpo}. Keep in mind that we are not primarily interested in providing means nor insights for hyperparameter optimization as a field. Rather, we use these problems as vehicle for the validation of our devised methods.

YAHPO Gym is a collection of hyperparameter optimization problems for classification tasks. It is constituted of a variety of different scenarios. A scenario refers to a set of problems which share the same hyperparameters, i.e., they are based on the same type of ML model. The different problem instances within a scenario are based on different OpenML datasets~\cite{OpenML2013}. The objective value of these problem instances is the misclassification error which needs to be minimized.

We excluded certain scenarios as these exhibited only continuous parameters (i.e., decision variables) and are of secondary interest in this work. Moreover, we excluded problems exceeding a dimensionality of $20$ due to the computational complexity. More on that in the next subsection.

A full enumeration of the considered scenarios is provided in Table~\ref{tab:yahpo}. It also details the size of the search space and how the number of continuous, integer, and categorical decision variables are distributed. In total, the considered subset of problems consists of $702$ MVPs.

It is important to note that the instances of a scenario are not necessarily related to shared problem difficulty or general landscape properties. For example, whether the fitness landscape is highly multi-modal or unimodal is contingent of not only a given scenario but also the dataset it is trained on. Hence, the grouping provided in Table~\ref{tab:yahpo} is for illustration purposes only and should not be interpreted as groupings of similar problems.
\begin{table}[htbp]
    \centering
    \caption{Overview of available scenarios in YAHPO Gym, the search space, the number of problem instances per scenario and information whether the search domain is hierarchical or not.}
    \label{tab:yahpo}
    \begin{tabular}{llll}
        \toprule
        Scenario & Search Space (cont, int, cat) & \# Instances & H \\
        \midrule
        rbv2\_glmnet & 3D: (2, 0, 1) & 115 & --\\
        rbv2\_rpart & 5D: (1, 3, 1) & 117 & --\\
        rbv2\_aknn & 6D: (0, 4, 2) & 118 & --\\
        rbv2\_svm & 6D: (3, 1, 2) & 106 & $\checkmark$ \\
        iaml\_ranger & 8D: (2, 3, 3) & 4 & $\checkmark$ \\
        rbv2\_ranger & 8D: (2, 3, 3) & 119 & $\checkmark$ \\
        iaml\_xgboost & 13D: (10, 2, 1) & 4 & $\checkmark$ \\
        rbv2\_xgboost & 14D: (10, 2, 2) & 119 & $\checkmark$ \\
        \bottomrule
    \end{tabular}
\end{table}

\subsection{Algorithm Portfolio}
For constructing our algorithm portfolio, we valued algorithms which are open-source, have been widely adopted, and work intrinsically different. The latter is of importance since we strive to get performance complementary algorithms.
We identified three different solvers with a promising performance and used random search as a baseline. The chosen algorithms are \texttt{SMAC3}~\cite{Lindauer_SMAC3_A_Versatile_2022}, \texttt{Optuna}~\cite{optuna_2019}, \texttt{pymoo}~\cite{pymoo}, and random search (RS).

\texttt{SMAC3} (SM) utilizes Bayesian optimization to solve optimization problems. The used surrogate model is, in contrast to regular Bayesian optimization approaches, a random forest and not a Gaussian process.

\texttt{Optuna} (OP) is an optimization framework with different algorithms. The considered one is operationally equivalent to SM but uses a tree-structured parzen estimator instead of a random forest. As these have been quite competitive and are functionally different surrogate models, we decided to include this variant into our algorithm portfolio.

On the other hand, \texttt{pymoo} (EA) is a framework based on evolutionary algorithms. The specific instance of \texttt{pymoo} we are using is a mixed-variable genetic algorithm where each type of decision variable has its own variation operator.

These algorithms are run with a budget of $100D$ where $D$ is the dimensionality of the MVP in question. Compared to other AAS studies, the allotted budget is fairly low. The cause for this is the substantial expense of computational resources used by the Bayesian methods especially, i.e., SM and OP. A single run takes several CPU hours up to a single day. This is consistent with the results of \cite{hutter13smac}. Furthermore, to gain any results of statistical significance, we run every algorithm $20$ times to get a realistic estimation of its average performance. Consequently, if we assumed that solving an MVP takes on average 4 CPU hours, which is a very conservative estimate, then we would spend computational resources of $468$ CPU days.

The performance of each algorithm is measured by the expected running time (ERT)~\cite{ert}, which is defined as follows:

\begin{equation}
    ERT = \frac{1}{s}\sum_j FE_j,
\end{equation}
where $j$ refers to one of the $20$ repetitions, $FE_j$ is the amount of function evaluations required to reach a target, and $s$ is the number of runs which successfully reached the target.

As we are minimizing the misclassification error in YAHPO Gym, we know that the theoretical optimum is zero. However, in practice this is rarely reached, especially for challenging problem instances. Hence, we determine the target by concatenating all optimizer traces over all the different repetitions for a given problem instance. This results in a vector which contains the objective values of each algorithm in each iteration. Based on this vector, we use the $0.01$-quantile of the objective values as a target. Thereby, we guarantee that at least one algorithm has solved each problem instance while also not making the problem instance trivial. If none of the $20$ repetitions reaches the target, we impute the ERT with the PAR10 score. This is in line with other works in that area (e.g., \cite{bischl2016aslib, KerschkeT2019AutomatedAlgorithm, PragerTWBK2020PerInstance}) and penalizes the maximal possible ERT by a factor of 10. 

Aggregated over all instances, the single best solver (SBS) of our algorithm portfolio is SM with an ERT of $13\,115$.
It is followed by EA with $38\,898$, OP with $96\,309$, and RS with $100\,674$. The ERT values are rounded to full numbers.

\begin{figure*}[t!]
    \centering
    \includegraphics[width=\textwidth]{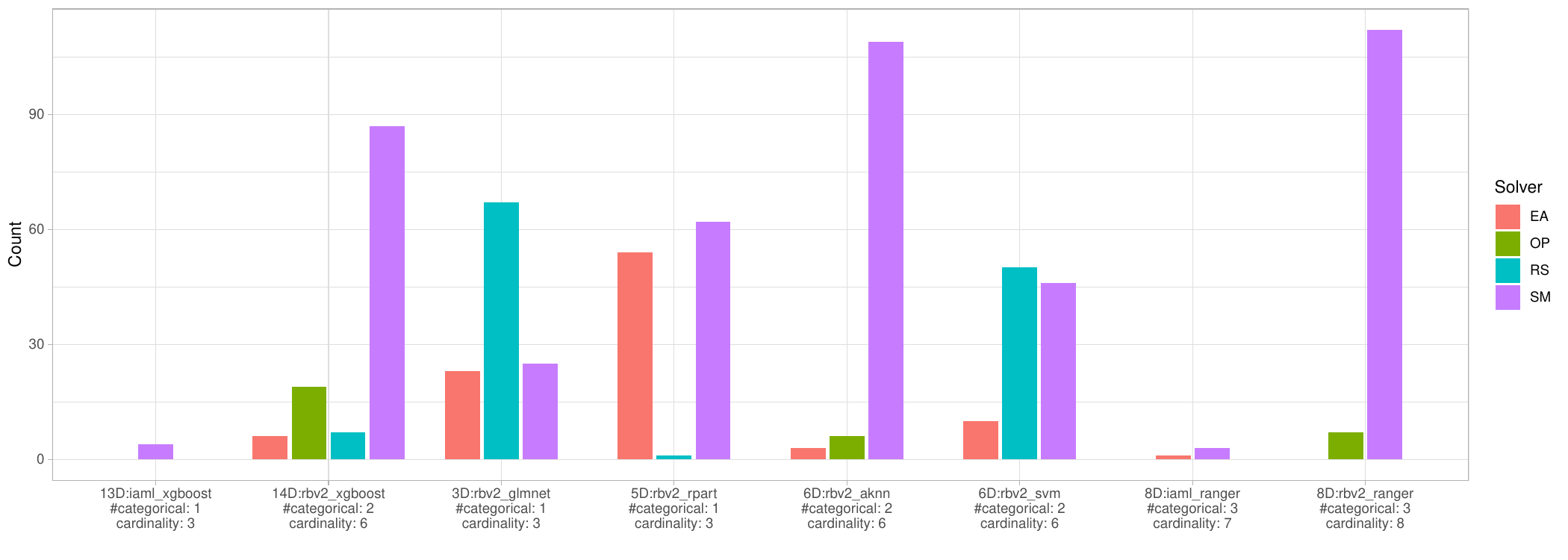}
    \caption{Frequency of best performing solver per scenario. The text on the x-axis states the problem dimension, the scenario name, the number of categorical variables and the sum of the cardinality of said categorical variables. The actual occurrences are as follows: SM $448$, RS $125$, EA $97$, OP $32$.}
    \label{fig:alg-distribution}
\end{figure*}

A more in-depth view is offered in Fig.~\ref{fig:alg-distribution}. Compartmentalized by scenario, it shows the frequency at which an algorithm performed best on a problem instance. The general superiority of SM is underlined further as we can observe that it is the best algorithm of our portfolio in the majority of cases. Nevertheless, in several cases one of the other three algorithms is more beneficial to use. For the scenario \texttt{rbv2\_glmnet} it is generally even better to rely on RS. This is not entirely surprising since RS has shown to be a strong baseline in hyperparameter optimization especially for problems of lower dimensionality~\cite{berg12randomsearch}.

To further gauge the potential of AAS, we contrast the SBS (i.e., SM) to the virtual best solver (VBS). In essence, the VBS represents a hypothetical algorithm selector which chooses the best algorithm out of our portfolio without fail. This VBS achieves an ERT of $444$, i.e., it requires almost $30$ times less budget than SM. Therefore, despite the superiority of SM, it is still worthwhile to pursue an AAS model. The actual composition of the VBS (i.e., how often a specific solver performed best on a problem instance) is as follows: SM $448$, RS $125$, EA $97$, OP $32$.

\begin{figure*}[t!]
    \centering
    \includegraphics[width=\textwidth]{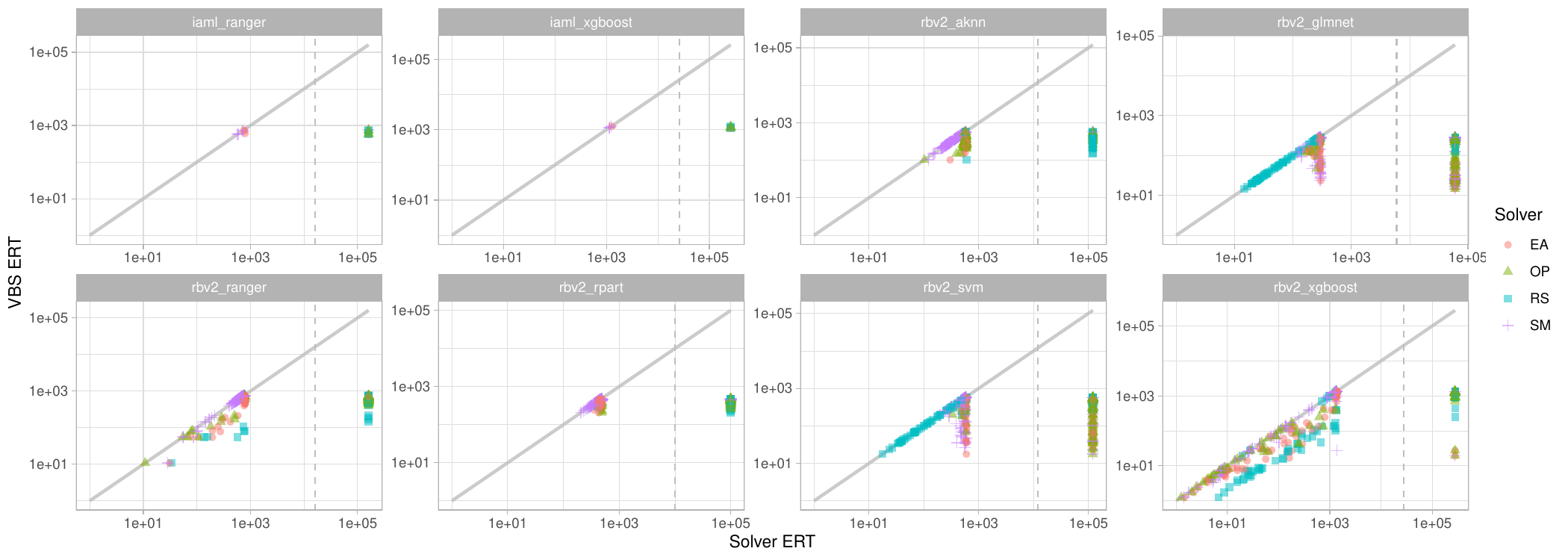}
    \caption{Performance comparison of each individual solver contrasted to the VBS on a log scale. When a solver has an equivalent performance to the VBS for a given problem instance, the points are located on the line diagonally separating each plot. The horizontal distance to that line quantifies how much worse a specific solver is compared to the VBS. The vertical dashed line represents worst possible ERT, where only a single repetition out of $20$ reaches the target.}
    \label{fig:gap}
\end{figure*}

At a first glance, the frequencies of the best performing solver contrasted against the actual ERT values seem contradictory. However, it shows that these solvers are truly performance complementary. Meaning, for a given problem instance an algorithm either solves it reliably well or not at all. This claim is supported by Fig.~\ref{fig:gap}. Here, we show how each individual solver compares against the VBS on a problem instance. It is evident that all of the four algorithms either perform consistently well for a given problem instance or do not reach the target in any repetitions at all. This is corroborated as there are no instances where a point is in close proximity to the dashed vertical line. This line indicates the worst possible ERT value for a given scenario. Only solvers with a majority of unsuccessful runs would be in the vicinity of this line.
Especially noteworthy, is the performance complementary of SM and RS. While SM dominates on the majority of scenarios, RS performs consistently well on the scenarios \texttt{rbv2\_svm} and \texttt{rbv2\_glmnet}.

\subsection{Exploratory Landscape Analysis Feature Generation}

\begin{figure*}[ht!]
    \centering
    \includegraphics[width=\textwidth]{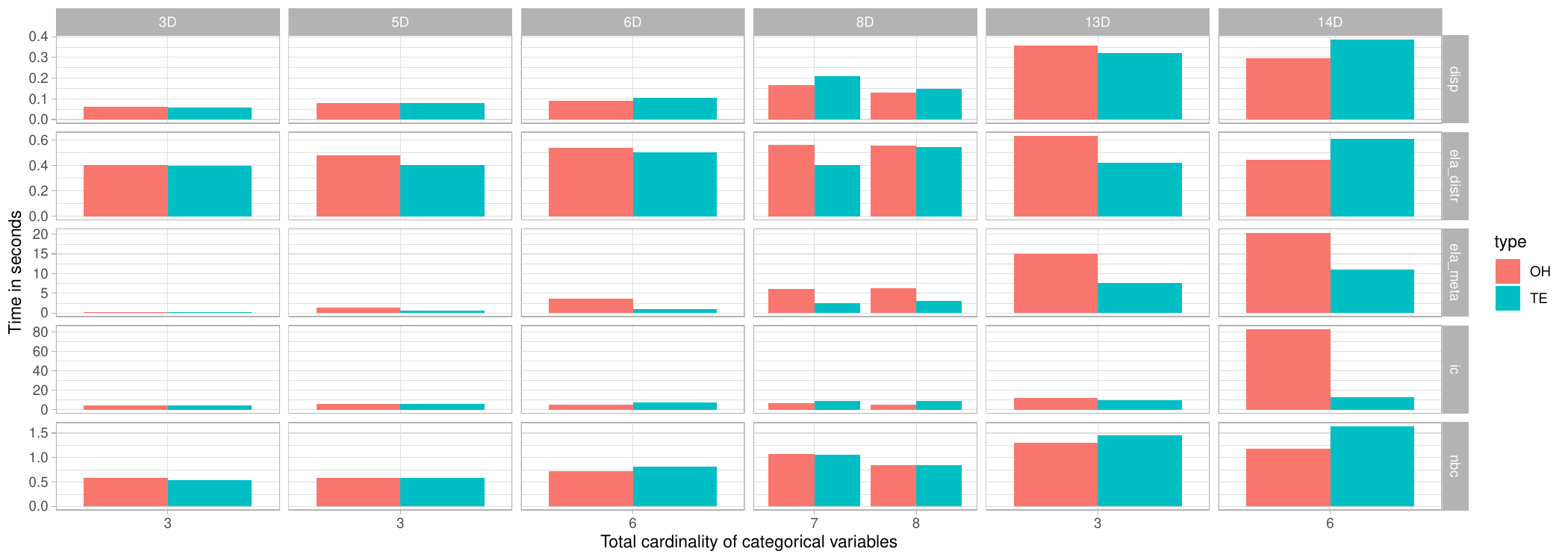}
    \caption{Computation time of ELA feature sets grouped by encoding and dimensionality of the problem. The x-axis depicts the total cardinality of the decision space whereas the y-axis shows the required time in seconds to calculate a respective feature set.  }
    \label{fig:ela-time}
\end{figure*}

\begin{figure*}
    \centering
    \includegraphics[width=\textwidth]{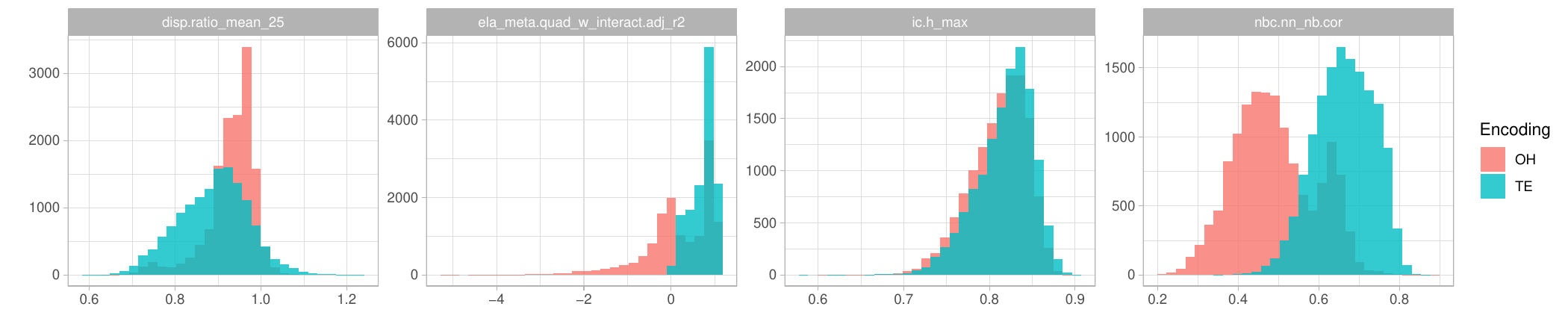}
    \caption{Histogram of four selected ELA feature values over all $702$ problem instances. The different colors represent the two encoding variants.}\label{fig:ela-hist}
\end{figure*}

\begin{figure*}[ht!]
    \centering
    \includegraphics[width=\textwidth]{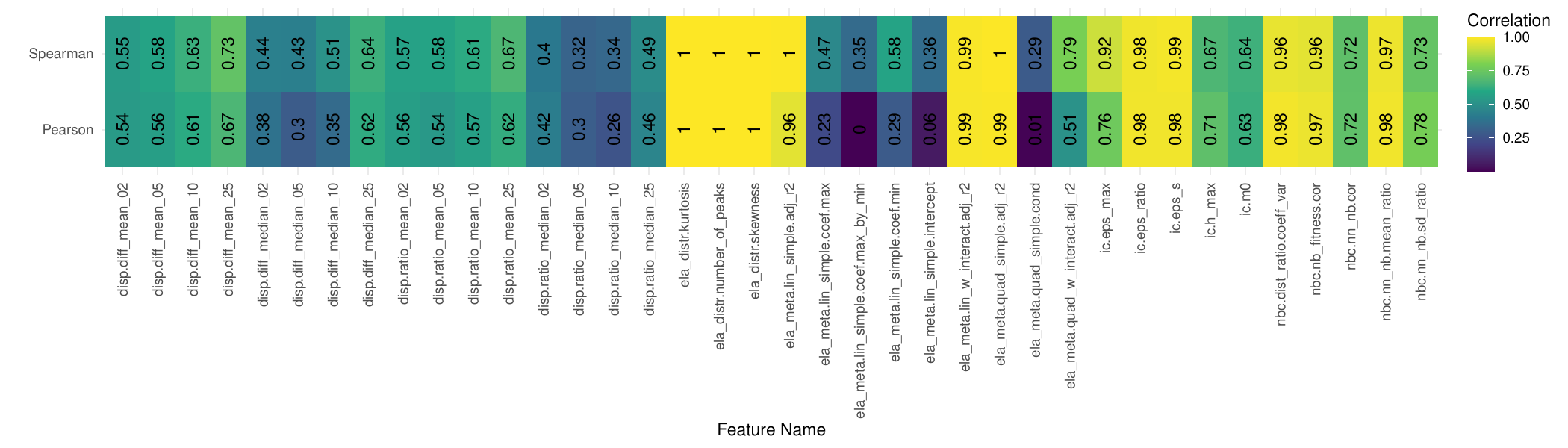}
    \caption{Bravais-Pearson and Spearman's rank correlation coefficient between the different encoding variants (TE and OH) for a given ELA feature over all considered problem instances.}
    \label{fig:ela-cor}
\end{figure*}

\begin{figure*}[ht!]
    \centering
     \begin{minipage}{0.49\textwidth}
        \includegraphics[width=\textwidth]{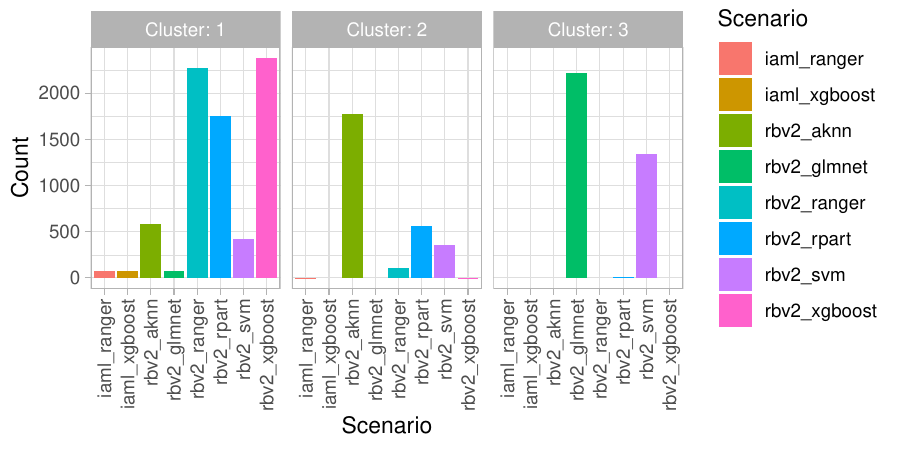}
        \caption{Frequency of problem instances of a scenario per cluster. Clusters are derived from ELA features based on TE.}
        \label{fig:bench-clust}
     \end{minipage}
    \begin{minipage}{0.49\textwidth}
        \includegraphics[width=\columnwidth]{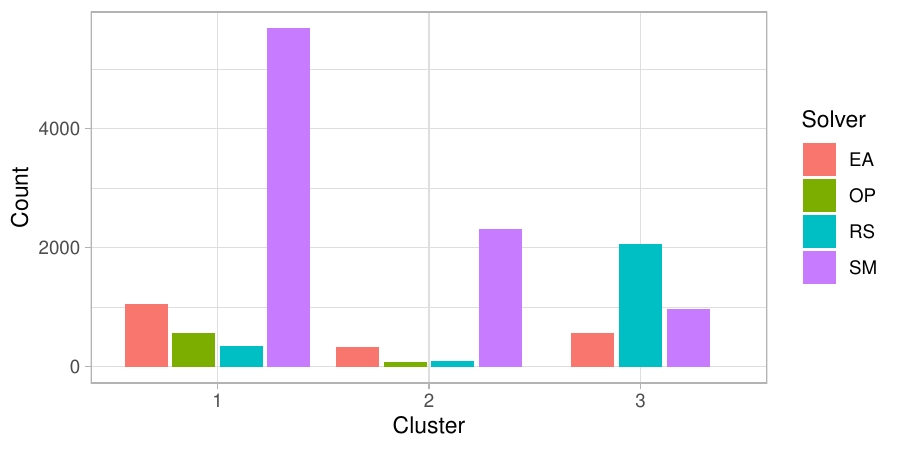}
        \caption{Frequency of the best-performing algorithm on a given problem instance divided by clusters. Clusters are derived from ELA features based on TE.}
        \label{fig:alg-clust}
    \end{minipage}
\end{figure*}


To enable our AAS model, we generate the relevant ELA features (cf. Section~\ref{sec:ela}) using the Python package \texttt{pflacco}~\cite{pflacco}. 
One of our earlier intuitions was that the cardinality of each categorical variable of an MVP might have an influence on the chosen encoding strategy. Meaning, we expect ELA features based on OH to be less descriptive than in the TE case. Hence, we complement the set of ELA features further with a feature which expresses the proportion of categorical variables in the decision space.

The initial sample is generated uniformly at random with a size of $50D$, where $D$ denotes the dimensionality of the problem at hand.
As we pointed out the detrimental effects of different ranges in the objective as well as the decision space, we normalize each decision variable and the objective value of a given sample between $[0, 1]$ independently.
Similar to the algorithmic case, we generate $20$ instances of ELA features per problem instance for each encoding variant to avoid stochastic interference to our findings.

As expected, the time required to calculate ELA features based on OH scales particularly bad with the dimensionality of the problem for some feature sets. This is especially true for the feature sets \texttt{ela\_meta} and \texttt{ic}. In the case of OH, the calculation of \texttt{ic} features takes on average $80$ seconds for the scenario \texttt{rbv2\_xgboost}. In comparison, TE only requires around $10$ seconds on average. This can be further inspected in Fig.~\ref{fig:ela-time}.
Interestingly enough, in case of the feature sets \texttt{disp} and \texttt{nbc} in the higher dimensions, TE requires slightly more time compared to OH. But this difference is measured in milliseconds and thus of lesser consequence. Especially so, since we measure the cost of feature computation in terms of actual spent function evaluation and not in CPU time.

In terms of values, there is no universal conclusion to be drawn. Fig.~\ref{fig:ela-hist} depicts the distribution of four features separated by the used encoding variant. The \texttt{disp} feature is based on distances to measure whether good fitness values are either localized or dispersed throughout the search space. Due to the increasing number of binary variables in OH, the distances between samples tend to be more extreme in contrast to TE, where a category can attain an arbitrary value in $[0, 1]$. One could argue, OH is less descriptive because it only offers a relatively small number of possible distances between observations. In contrast, some features like \texttt{ic.h\_max} have little difference between the two encoding variants. Other features such as \texttt{ela\_meta.quad\_w\_interact.adj\_r2} and \texttt{nbc.nn\_nb.cor} resemble a bimodal distribution with OH encoding whereas TE is unimodal. 

A more holistic picture is given in Fig.~\ref{fig:ela-cor} which depicts the Bravais-Pearson and Spearman's rank correlation coefficient between the different encoding variants for a given ELA feature. We can observe a low correlation for the entire \texttt{disp} feature set and features of \texttt{ela\_meta} which are related to the coefficients of the linear or quadratic model. 
These low correlation coefficients can be traced back to the dimensionality of our initial sample $S$ \textit{after} transformation. TE does not affect the dimensionality of $S$ whereas OH increases it depending on the number of categorical variables and their respective cardinality. Hence, distances generally tend to be larger in the OH case which affects the \texttt{disp} feature set. In the case of \texttt{ela\_meta}, the growing dimensionality of $S$ raises the number of model coefficients. Consequently, the probability for larger or smaller model coefficients rises which in turn directly influences feature such as \texttt{ela\_meta.lin\_simple.coef.max}.
 
Based on the aforementioned analyses, we cannot determine which of the encoding is generally or situationally superior. We see some drawbacks for the OH variant, but this is largely limited to the computation time required as well the diversity of some isolated ELA features. We expect that potential issues will reveal themselves on problems of higher dimensionality with a substantial amount of categorical variables.

To further deepen our understanding of ELA features in the context of MVPs, we perform a cluster analysis to evaluate whether meaningful groupings of the benchmark problems can be achieved based on ELA features. As pointed out in the subsection related to the benchmark functions, the current division into scenarios does not necessarily hold any meaning. We evaluated the cluster algorithms DBSCAN~\cite{dbscan}, $k$-Means, and several agglomerative hierarchical approaches~\cite{hastie2009elements}. The most meaningful clustering result is achieved with hierarchical clustering using the ward linkage method. Based on close inspection of the respective dendrogram, this results in three clusters. This is based on the TE variant. As we could not find a satisfactory low dimensional representation of the original dataset, we decided to exclude a figure pertaining this.

Fig.~\ref{fig:bench-clust} and Fig.~\ref{fig:alg-clust} show the distribution of scenarios and the distribution of best-performing algorithms per cluster respectively. While we hypothesized that problem instances belonging to a specific scenario not necessarily share the same landscape characteristics, Fig.~\ref{fig:bench-clust} contradicts this to a lesser extent. Cluster 1 comprises a multitude of problems instances belonging to different scenarios. The majority of these problem instances belong to scenarios on the middle to higher end in terms of dimensionality. Cluster 2 includes the majority of problem instances of the scenario \texttt{rbv2\_aknn} as well as some instances of a few other scenarios. It does not comprise any problem instances with over 8 dimensions and only a few at that. Cluster 3 is mainly constituted by the two scenarios \texttt{rbv2\_glmnet} and \texttt{rbv2\_svm}. This is of particular interest as these problem instances are largely absent in the other two clusters. We could not find any statistical association to the quantity of categorical decision variables, to the total cardinality of these, or to the presence of hierarchical dependencies between decision variables.

Nevertheless, we can augment the drawn conclusion by digesting the information of Fig.~\ref{fig:alg-clust}. This figure depicts the distribution of the best-performing algorithms (similar as in Fig.~\ref{fig:alg-distribution}) divided by clusters. While there is little difference between cluster 1 and 2 from an algorithm ranking perspective, cluster 3 contains mostly problem instances where RS is superior. This underlines the differences between cluster 3 compared to cluster 1 and 2.

\subsection{Automated Algorithm Selection}
We model our AAS scenario as a multi-class classification task. Meaning, for a given problem instance, we use the ELA features as input for an ML model. The class label in that case is the best performing algorithm on that particular problem instance, i.e., the solver with the lowest ERT. Note that our dataset comprises $702 \cdot 20 = 14\,040$ observations for each encoding variant because we generated ELA features $20$ times per problem instance (of which we have $702$). The class label for these $20$ repetitions however remains the same.

As options for ML models, we consider random forests, support vector machines and gradient boosting machines~\cite{hastie2009elements}. These are implemented in the Python package \texttt{scikit-learn}~\cite{scikit-learn}.

Previous studies pointed out that feature selection can greatly influence the performance of AAS models~\cite{KerschkeT2019AutomatedAlgorithm}. We decided to employ this as well. The selected approach in our implementation is commonly referred to as `greedy forward-backward selection'. Commencing with an initial feature set devoid of any features, this selection technique adopts a greedy approach by iteratively incorporating and/or eliminating features, which is contingent upon the enhancement of the model's performance in subsequent iterations. To perform the described feature selection method, we utilize the Python package \texttt{mlxtend}~\cite{raschkas2018mlxtend}.
With a set of ELA features of size $40$, feature selection itself is computationally expensive. Hence, we allocate only limited resources to tuning the respective model parameters. Meaning, we train each model with its default configuration in addition to $4$ randomly sampled configurations.

The model evaluation is performed by using $10$-fold cross validation, where all repetitions of any given problem instance belong to a singular fold. With that, we avoid that repetitions of a problem instance are distributed across multiple folds. This would have inevitably led to circumstances where a problem instance is trained and simultaneously tested on.

\begin{table*}[ht!]
\centering
\caption{Cells hold the arithmetic mean of the ERT value for the respective settings. Values with a gray background indicate the type of solver which performed best for a given setting. The column `Gap Closure' represents the closure of the gap between the SBS and VBS by the TE model as percentage.}
\label{tab:aas-performance}
\begin{tabular}{ll|c|c|cc|c}
  \toprule
\textbf{Scenario} & \textbf{\# Instances} & \textbf{VBS} & \textbf{SBS} & \textbf{OH} & \textbf{TE} & \textbf{Gap Closure} \\ 
  \midrule
rbv2\_glmnet & 2\,300 & 165.46 & 28\,312.86 & 11\,590.63 & {\cellcolor{gray!50}11\,437.34} & 59.95 \\ 
   \midrule rbv2\_rpart & 2\,340 & 412.47 & 13\,182.87 & 3\,648.72 & {\cellcolor{gray!50}2\,755.18} & 81.66 \\ 
   \midrule rbv2\_aknn & 2\,360 & 354.90 & 2\,380.31 & 2\,730.56 & {\cellcolor{gray!50}2\,375.83} & 0.22 \\ 
   \midrule rbv2\_svm & 2\,120 & 339.59 & 25\,297.37 & 15\,310.14 & {\cellcolor{gray!50}12\,498.84} & 51.28 \\ 
   \midrule iaml\_ranger &  80 & 659.55 & 40\,471.21 & 38\,880.63 & {\cellcolor{gray!50}32\,908.88} & 19.00 \\ 
   \midrule rbv2\_ranger & 2380 & 585.70 & {\cellcolor{gray!50}587.45} & 1\,254.72 & 986.87 & - \\ 
   \midrule iaml\_xgboost &  80 & 1\,170.40 & {\cellcolor{gray!50}1\,170.40} & 1\,820.40 & 1\,820.40 & - \\ 
   \midrule rbv2\_xgboost & 2\,380 & 754.12 & 10\,162.03 & 5\,804.02 & {\cellcolor{gray!50}4\,978.53} & 55.10 \\ 
   \midrule \midrule Cluster 1 & 7\,649 & 573.00 & 8\,902.08 & 5\,272.84 & {\cellcolor{gray!50}4\,019.43} & 58.62 \\ 
   \midrule  Cluster 2 & 2\,811 & 380.00 & 4\,680.77 & 2\,973.21 & {\cellcolor{gray!50}2\,631.63} & 47.65 \\ 
   \midrule Cluster 3 & 3\,580 & 219.91 & 28\,737.67 & 12\,699.54 & {\cellcolor{gray!50}12\,204.30} & 57.98 \\ 
   \midrule \midrule All & 14\,040 & 444.33 & 13\,114.71 & 6\,706.12 & {\cellcolor{gray!50}5\,828.60} & 57.51 \\ 
   \midrule  \bottomrule
\end{tabular}
\end{table*}

\begin{table}[ht!]
\centering
\caption{Listings of the identified feature subsets for both models. The order of features is based on the results of the feature selection method. Cells marked in gray highlight ELA features which are only present in one model. The feature \texttt{nbc.nb\_fitness.cor} has a darker hue to emphasize that it is the only case where no equivalent feature is present in the other model.}
\label{tab:sffs-features}
\begin{tabular}{ll}
  \toprule
\textbf{Model based on TE} & \textbf{Model based on OH} \\ 
  \midrule
ela\_distr.number\_of\_peaks & ela\_distr.number\_of\_peaks \\ 
   \midrule ela\_meta.lin\_simple.intercept & ela\_meta.lin\_simple.intercept \\ 
   \midrule ela\_meta.lin\_simple.coef.min & ela\_meta.lin\_simple.coef.min \\ 
   \midrule ela\_meta.quad\_simple.adj\_r2 & {\cellcolor{gray!20}ela\_meta.lin\_w\_interact.adj\_r2} \\ 
   \midrule ela\_meta.quad\_simple.cond & ela\_meta.quad\_simple.adj\_r2 \\ 
   \midrule {\cellcolor{gray!20}ela\_meta.quad\_w\_interact.adj\_r2} & ela\_meta.quad\_simple.cond \\ 
   \midrule {\cellcolor{gray!50}nbc.nb\_fitness.cor} & disp.diff\_mean\_02 \\ 
   \midrule disp.diff\_mean\_02 & disp.diff\_mean\_05 \\ 
   \midrule disp.diff\_mean\_05 & {\cellcolor{gray!20}disp.diff\_mean\_10} \\ 
   \midrule {\cellcolor{gray!20}disp.diff\_median\_02} & {\cellcolor{gray!20}disp.diff\_mean\_25} \\ 
   \midrule {\cellcolor{gray!20}disp.diff\_median\_05} & ic.h\_max \\ 
   \midrule {\cellcolor{gray!20}disp.diff\_median\_10} & ic.eps\_s \\ 
   \midrule ic.h\_max & - \\ 
   \midrule ic.eps\_s & - \\ 
   \midrule  \bottomrule
\end{tabular}
\end{table}

The best performing model for both encoding variants is a random forest. In the case of OH, the feature selection efforts yielded a model based on $12$ features whereas in the case of TE it results in $14$. The two feature sets largely overlap as can be seen in Table~\ref{tab:sffs-features}. Since the employed feature selection method is a greedy approach, we can attribute some form of importance to the order of the presented features. Both models start with incorporating the feature \texttt{ela\_distr.number\_of\_peaks}. This is followed by an array of \texttt{ela\_meta} features. In general, the feature sets \texttt{ela\_meta} and \texttt{disp} dominate the other feature sets in terms of frequency. Features of the set information content are last in both models.
While there exist some differences between the subset of features between both models, they are mostly inconsequential. For example, TE uses the $R^2$ value of a quadratic model with interaction whereas OH incorporates the same metric for a linear model. This also applies to discrepancies related to the dispersion feature set where metrics based on the median are replaced with the arithmetic mean.

The only notable difference is that TE also utilizes a singular feature of the nearest better clustering feature set. In OH, this feature set is eliminated entirely. Moreover, we find these results interesting as the clustering analysis based on ELA features hinted that dimensionality plays a crucial role in dividing the problem instances. Yet, the dimension (also an input variable for the AAS model) does not exist in any subset of features. We assume that this information is encapsulated in one or more features shown in Table~\ref{tab:sffs-features}.

The actual performance of our AAS model achieves an accuracy of $0.712$. However, the accuracy is only of secondary concern as we are more interested in how well the actual predictions would perform on a problem instance.
Therefore, we measure how well the predictions of the AAS models perform in terms of ERT values.
An aggregation of the performance on different levels is detailed in Table~\ref{tab:aas-performance}. Each cell represents the ERT values for a given scenario or cluster and a given optimization approach. These optimization approaches are the lower bound (VBS), baseline (SBS), and the two AAS models based on the two different encoding variants (OH and TE). Cells which are colored in gray highlight the best performing solver out of the SBS, TE and OH model. As feature computation comes at a cost - namely the size of the initial sample - we include the respective costs (measured in function evaluations) in the performance of the OH and TE AAS models.

Overall, we can conclude that TE is superior to OH by a considerable margin. This margin is in some cases very small (e.g., in the scenarios \texttt{rbv2\_glmnet} and \texttt{iaml\_xgboost}) but more often it adds an additional cost of hundreds to the ERT.
The juxtaposition of TE and the SBS provides a more heterogeneous picture. In the majority of cases, our AAS model based on TE is largely superior to the SBS. In the rare instances where this is not the circumstance, this can be traced back to the general superiority of the SBS for that particular subset of problems. This pertains to the scenarios \texttt{rbv2\_ranger} and \texttt{iaml\_xgboost}. Consulting Fig.~\ref{fig:alg-distribution} which showed the distribution of best performing algorithm per scenario, we can see that the SBS was either always the best choice or at least in over $90\%$ of the problem instances. For these type of problem instances, the additional costs induced by feature calculation is too expensive to be competitive. The same case can be made for the scenario \texttt{rbv2\_aknn}. While the TE model is better, the difference is minuscule bordering on the edge where feature calculation is too costly again. The perspective derived from the clustering results does not provide further insight. Here, our AAS model based on TE consistently performs well and closes the gap between the SBS and VBS by almost half.

Aggregated overall problem instances, the TE approach achieves an ERT value of $5\,829$ and the SBS remains at $13\,115$. Consequently, our devised model is able to close the SBS-VBS gap by around $57.5\%$. This is a very conservative estimate of the performance gain to be had. We are confident that these satisfying results can be improved in various manners without inducing additional costs in terms of function evaluations. For instance, both SM and OP are Bayesian optimization methods which require an initial sample as well in order to train their respective surrogate model. Admittedly, this is typically smaller than the initial sample of ELA yet it can still be utilized. This would accelerate the subsequent search for the target of a problem instance. Similarly, the initial design can be used as an initial population for the solver EA.
The case for RS is even more straightforward as the current sampling strategy of the ELA computation is based on uniform random sampling, i.e., we could stop the solving process altogether when the sample for the ELA computation already reached the target.

\section{Conclusion and Outlook}\label{sec:conclusion}
A myriad of problem instances present in academia and the real-world are constituted of various types of decision variables. These MVP often impose an additional layer of complexity for optimization algorithms as well as the analysis of their fitness landscape.
Within this work, we propose a preprocessing scheme to enable the calculation of existing fitness landscape features in the domain of MVP. 
This preprocessing scheme should be employed prior to ELA feature computation. Each component of this approach addresses a particular challenge introduced by mixed-variable search spaces. The primary focus of this work is the evaluation of two different mechanisms to transform discrete decision variables without ordinal structure to a numerical representation and their subsequent application to fitness landscape analysis. These selected encoding variants are one-hot encoding and target encoding.

We evaluate the merit of our devised approach by applying it on a set of hyperparameter optimization problems. The resulting landscape features, based on the different encoding variants, are analyzed rigorously. Features originating from these encoding variants do not paint a clear picture, i.e., some features are highly correlated and are similarly distributed, yet other features exhibit clear disparities between one-hot and target encoding. In general, we observe that features using target encoding are faster to compute by a substantial margin. This can be measured in seconds and up to minutes.

Our preprocessing scheme is put to the test in an algorithm selection study. In there, we use the aforementioned set of problem instances as well as a set of performance complementary algorithms. The objective is to construct an ML model which is able to automatically select an appropriate solver for a given problem instance. Our endeavor in that regard is successful as we are able to close the gap between the SBS and VBS by around $57.5\%$. The most influential features stem from the feature sets \texttt{ela\_meta} and \texttt{disp}. The encoding TE is superior in terms of performance compared to OH by a considerable margin. Given TE's generally faster computation as well as its behavior to result into a single scalar independent of the cardinality of a categorical variable, TE is the recommended mechanism to employ in future research.

However, there still remain large areas unexplored. One of which pertains to the sampling strategy for mixed-variable search spaces. In the continuous domain, the switch from random samples to space filling designs has shown to improve the discriminating power of ELA features~\cite{renau2020sampling}. Thus, we deem it worthwhile to invest more resources in that particular area in the future.

A second imminent concern of ours is the development of more sophisticated strategies to deal with hierarchical relationships between decision variables. While our proposed preprocessing scheme is able deal with them sufficiently, we hypothesize that these structures can be leveraged in a tailored manner. Our current strategy is to impute the indeterminate value of an `inactive' decision variable with a random value out of its domain \textit{a single time}. Another option could be the creation of multiple random values as these come at no cost. Thereby, we would capture these plateau shape areas in more detail. Yet, this has to be balanced as this ultimately would lead to oversampling of certain areas compared to regions with sparse information. 

While these two options for future research pertain to direct improvements of our devised methods, other areas which require further examination are the general relationship between ELA features and high-level properties of MVPs. This can take form as an ablation studies to measure for example which feature sets can be used to predict the presence of funnel structures in MVPs.
Nevertheless, given our promising findings and the various avenues to further improve our work, we expect a rise in interest in fitness landscape analysis in general and especially in the domain of mixed-variable optimization.

\bibliographystyle{IEEEtran}
\bibliography{99_references}


 




\begin{IEEEbiographynophoto}{Raphael Patrick Prager}
received his B.Sc. and M.Sc. degree in Information Systems from the University of M\"unster where he currently also pursues a Ph.D. degree. He worked for three years as a data scientist at the chemical company BASF where his contributions focused around the acceleration of product development via optimization and machine learning techniques. His research primarily pertains to fitness landscape analysis, automated algorithm selection, and black-box optimization.
\end{IEEEbiographynophoto}

\begin{IEEEbiographynophoto}{Heike Trautmann}
Heike Trautmann received her PhD and Habiltation at TU Dortmund University, Germany. After 10 years as professor of Data Science: Statistics and Optimization at the University of Münster, Germany, she is currently Professor of Machine Learning and Optimisation at the Department of Computer Science, University of Paderborn, Germany and also Guest Professor of Data Science at the University of Twente, NL. Her research mainly focuses on Trustworthy AI, Data Science, Automated Algorithm Selection and Configuration, Data Stream Mining and Evolutionary Optimization. She is member of the ACM SIGEVO executive committee and associate editor of the IEEE Transactions on Evolutionary Computation (TEVC) as well as the Evolutionary Computation Journal (ECJ).
\end{IEEEbiographynophoto}

\vfill

\end{document}